 \documentclass[letterpaper, 10 pt, conference]{ieeeconf}  

\IEEEoverridecommandlockouts

\usepackage{graphics} 
\usepackage{times} 
\usepackage{amsmath} 
\usepackage{amssymb}  
\usepackage{soul}
\usepackage{multirow}
\usepackage{booktabs}
\usepackage{relsize}
\usepackage{xspace}
\usepackage{bbold}
\usepackage{subfig}
\usepackage{listings}
\usepackage{bm}
\usepackage{bbm}
\usepackage{mathtools}
\usepackage{float}
\usepackage{soul}
\usepackage{textcomp}
\usepackage{gensymb}
\usepackage{tabularx}
\usepackage{makecell}
\usepackage{multirow}
\usepackage{hyperref}
\usepackage{booktabs}

\newcommand{\zheyu}[1]{\textcolor{blue}{}\textcolor{black}{#1}}

\usepackage[color=todocolor, colorinlistoftodos]{todonotes}

\DeclareMathOperator{\se3}{\mathbf{SE(3)}}

\newcommand{\R}{\mathbb{R}}
\newcommand*{\eye}{\mathbf{1}}
\newcommand*{\zero}{\mathbf{0}}

\newcommand*{\C}[1]{\mathcal{#1}}
\newcommand*{\lyap}{\mathcal{V}}

\newcommand*{\pose}[2]{\prescript{#2}{}{\bm{X}}_{#1}}
\newcommand*{\trans}[2]{\prescript{#2}{}{\bm{p}}_{#1}}

\newcommand*{\rot}[2]{\prescript{#2}{}{\bm{R}}_{#1}}

\newcommand{\textjava}[1]{{\lstset{basicstyle=\ttfamily}\lstinline@#1@}}
\newcommand{\textjavafn}[1]{{\lstset{basicstyle=\footnotesize\ttfamily}\lstinline@#1@}}
\usepackage{setspace}
\usepackage{ifthen}

\long\def\sfootnote[#1]#2{\begingroup%
\def\thefootnote{\fnsymbol{footnote}}\footnote[#1]{#2}\endgroup}
\newcommand{\SE}{\mathrm{SE}}
\newcommand{\SO}{\mathrm{SO}}
\renewcommand{\se}{\mathfrak{se}}
\newcommand{\ie}{i.e., }

\newcommand{\ignore}[1]{}

\DeclareMathOperator*{\argmin}{arg\,min}
\renewcommand{\baselinestretch}{0.99}

\title{\LARGE \bf
LyRN (Lyapunov Reaching Network):
A Real-Time Closed Loop approach from Monocular Vision
}
\author{Zheyu Zhuang$^{1}$, Xin Yu$^{1}$, Robert Mahony$^{1}$
\thanks{This research was supported by the Australian Research Council
through the ``Australian Centre of Excellence for Robotic Vision'' CE140100016.}%
\thanks{$^1$ Zheyu Zhuang, Xin Yu, Robert Mahony are with ``Australian Centre for Robotic Vision'', Research School of Engineering, The Australian National University, Canberra ACT, 2601, Australia. {\tt\small first.last@anu.edu.au}}
}%

\begin{document}

\maketitle
\thispagestyle{empty}
\pagestyle{empty}

\begin{abstract}
We propose a closed-loop, multi-instance control algorithm for visually guided reaching based on novel learning principles.
A control Lyapunov function methodology is used to design a reaching action for a complex multi-instance task in the case where full state information (poses of all potential reaching points) is available.
The proposed algorithm uses monocular vision and manipulator joint angles as the input to a deep convolution neural network to predict the value of the control Lyapunov function (cLf) and corresponding velocity control.
The resulting network output is used in real-time as visual control for the grasping task with the multi-instance capability emerging naturally from the design of the control Lyapunov function.

We demonstrate the proposed algorithm grasping mugs (textureless and symmetric objects) on a table-top from an over-the-shoulder monocular RGB camera.
The manipulator dynamically converges to the best-suited target among multiple identical instances from any random initial pose within the workspace.
The system \zheyu{trained with only simulated data} is able to achieve 90.3\% grasp success rate \zheyu{in the  real-world experiments} with up to 85Hz
closed-loop control on one GTX 1080Ti GPU and significantly outperforms a Pose-Based-Visual-Servo (PBVS) grasping system adapted from a state-of-the-art single shot RGB 6D pose estimation algorithm.
A key contribution of the paper is the inclusion of a first-order differential constraint associated with the cLf as a regularisation term during learning, and we provide evidence that this leads to more robust and reliable reaching/grasping performance than vanilla regression on general control inputs.

\end{abstract}

\section{INTRODUCTION}


Visual servo control is an established research field in robotics~\cite{kemp2007challenges}.
Historically, visual servoing algorithms are categorised into either pose-based visual servoing (PBVS) or image-based visual servoing (IBVS)~\cite{hutchinson1996tutorial,corke1993visual}.
Pose-based approaches use image-based pose estimation techniques coupled to a classical pose based robot controller.
Image-based methods, on the other hand, generate control signals directly from an error function derived from input images~\cite{corke2000real}.
Classical IBVS algorithms are generally considered superior to classical PBVS methods since they are robust to model errors and camera calibration.

Recently, deep learning pose estimation algorithms have been shown to achieve superior performance in comparison to classical visual pose estimation methods.
Monocular vision pose estimation algorithms~\cite{Rad:2017et, Tekin:2017wk, brachmann2014learning, brachmann2016uncertainty} utilise deep neural networks to extract image features and establish 2D-3D point correspondences, thus obtaining 6 degree of freedom (DoF) object poses by solving a PnP problem.
PoseCNN~\cite{xiang2017posecnn} firstly predicts a semantic label for each object and then estimates the centers and rotations of objects in an image.
Tekin et al.~\cite{tekin2018real} present a real-time multi-class single-shot 6DoF pose estimation method from RGB images.
Their method exploits a convolutional network (CNN) to regress 2D projections of the 3D bounding box of an object and then estimates object poses by solving a PnP problem.
Object pose estimation algorithms~\cite{xiang2017posecnn, peng2019pvnet} are
designed to find class-specific objects in an image.
Such architectures assume one instance per class in any given image~\cite{hinterstoisser2012model,xiang2017posecnn} and do not generalise well to multi-instance grasping tasks common in robotics.
Detection based pose estimation approaches~\cite{kehl2017ssd,tekin2018real, sundermeyer2018implicit} can address multiple instance cases, but are subject to false positive detection.
Coupling deep learned pose estimation with a classical robotic controller is an example of a modern PBVS control architecture.

The applicability of deep learning to IBVS control is not as clear.
There is an impressive body of work on visual grasping, where the goal is to find a good grasp point on an arbitrary object.
GG-CNN~\cite{morrison2020GGCNN} and Tossingbot~\cite{zeng2019tossingbot} choose an object based on pixel-wise grasping probabilities estimated from vision input, while reinforcement learning based grasping algorithms~\cite{levine2016end, Levine:2017cu, kalashnikov2018qt} select an action that maximises the grasping rewards without taking the class information of objects into account.
The problem of reaching and grasping, where the goal is to reach for and grasp a specific type of object, is a qualitatively different problem requiring instance segmentation to choose a goal from possibly multi-instances, but then admitting priors on how to grasp an object.
Zhang~\textit{et al.}~\cite{Fangyi2018} achieve closed-loop reaching towards one unique target in a clutter from monocular RGB images by learning a visuo-motor policy from a pose-based controller.

In this paper, we propose a YOLO-like~\cite{redmon2017yolo9000} single-shot CNN architecture that takes a monocular RGB image and current manipulator joint angles as input to directly compute the joint velocity input for a multi-instance reaching and grasping task.
The control algorithm that we learn is based on classical control Lyapunov function design and a key innovation of the paper is to learn the value of the Lyapunov function along with the joint velocity control and regularise the learning task with a first order differential Lyapunov decrease constraint implemented through a Siamese network architecture.
This innovation is compared, and shown to be superior, to a vanilla deep learning implementation where only the joint velocity control is learned by the network.
In our multi-instance reaching/grasping task, there are multiple identical target candidates in the workspace, and each candidate corresponds to a cLf value and corresponding joint velocity.
Unlike the classical `decide and servo' paradigm, we implement the control associated with the minimum cLf prediction in each frame.
The resulting control adapts to reach for the `closest' target as perceived by the vision system and will function effectively even in a dynamically changing environment.
The contributions of the paper are:
\begin{itemize}
    \item To demonstrate real-time, closed loop, fast multi-instance visual reaching and grasping in cluttered and dynamic environments from a first-person RGB camera.
    \item To demonstrate the proposed algorithm is more robust to the sim-to-real domain gap compared to a PBVS system adapting from a state-of-the-art monocular RGB pose-estimation algorithm.
    \item To show that correlating a control Lyapunov function with the control inputs via a differential constraint \zheyu{in training CNNs} leads to better reliability and performance in comparison to vanilla regression.
\end{itemize}

Section \ref{sec:cLf_design} presents the controller design.
Section \ref{sec:III:learning} describes the network architectures, loss functions and a regularisation term to impose the inherent differential constraint in Lyapunov controller design.
The implementation details are included in Section \ref{sec:implementation}.
Section \ref{sec:results} demonstrates the experimental study that evaluates the real-world performance of our method against a PBVS baseline using from a state-of-the-art RGB based pose-estimation algorithm and a vanilla regression baseline.
\section{Controller Design}
\label{sec:cLf_design}

This section presents the formulation of the control Lyapunov function for a reaching task.
\subsection{Symmetry-aware Control Lyapunov Function}
The 6 Degrees of Freedom (DoF) pose of a target and an end-effector are represented by elements of the Special Euclidean Group $\SE(3)$.
Denote the pose of a general frame $\{A\}$ with respect to another frame $\{B\}$ as $\pose{A}{B}$, and its rotation matrix and translation vector as $\rot{A}{B} \in \SO(3)$ and $\trans{A}{B}\in \R^3$ respectively.
The left superscript is omitted if the pose is defined with respect to the world reference frame.
Denote the end-effector frame as $\{H\}$ and the target frame as $\{G\}$.
The absolute end-effector pose $\pose{H}{} = \pose{H}{}(\bm{\theta})$ is a function of joint angles $\bm{\theta}\in\R^{6\times1}$, i.e., the forward kinematics model of the manipulator.\par
A cLf for a reaching task is a continuously differentiable scalar-valued positive-definite function $\lyap(\bm{\theta})$ of the joint angles that is zero only at the joint coordinates for the desired pose.
The aim of a reaching task for a goal $\{G\}$ is to drive the end-effector $\pose{H}{}(\bm{\theta})$ to the goal $\pose{G}{}$. In this work, we formulate a control Lyapunov function based on $\SE(3)$ object poses. We start from defining a cLf as
\[
    \frac{1}{2}\|\pose{H}{}-\pose{G}{}\|^2_F,
\]
where $\| \cdot \|_F$ denotes the Frobenius norm. However, this cLf disregards reflective and rotational symmetry of the targets. 
To address this issue, we define
\begin{multline}
	\label{eq:cLf_complex}
	\lyap := \frac{1}{2}\|\pose{H}{}-\pose{i}{}\pose{G}{i}^*\|^2_F,\: \text{for } \{\pose{G}{i}^*\vert\rot{G}{i}^*= \\ \argmin_{\rot{G}{i}}\|\rot{H}{}-\rot{i}{}\rot{G}{i}\|,\: \trans{G}{i}=\zero_{3\times1}\},
\end{multline}
where $\pose{i}{}$ and $\pose{G}{i}^*$ denote a non-optimal target pose and the optimal transformation with respect to the target that minimises the cLf respectively.
Define $\pose{G}{}^* = \pose{i}{}\pose{G}{i}^*$.
In this work, where we consider grasping a mug, the number of rotational symmetry axes is 1, the remaining axes can be configured as reflective symmetry axes.
The  body fixed frame attached to target objects is aligned to the rotational symmetry axis.

\begin{figure*}
	\centering
	\subfloat[]{
		\includegraphics[width=0.9\linewidth]{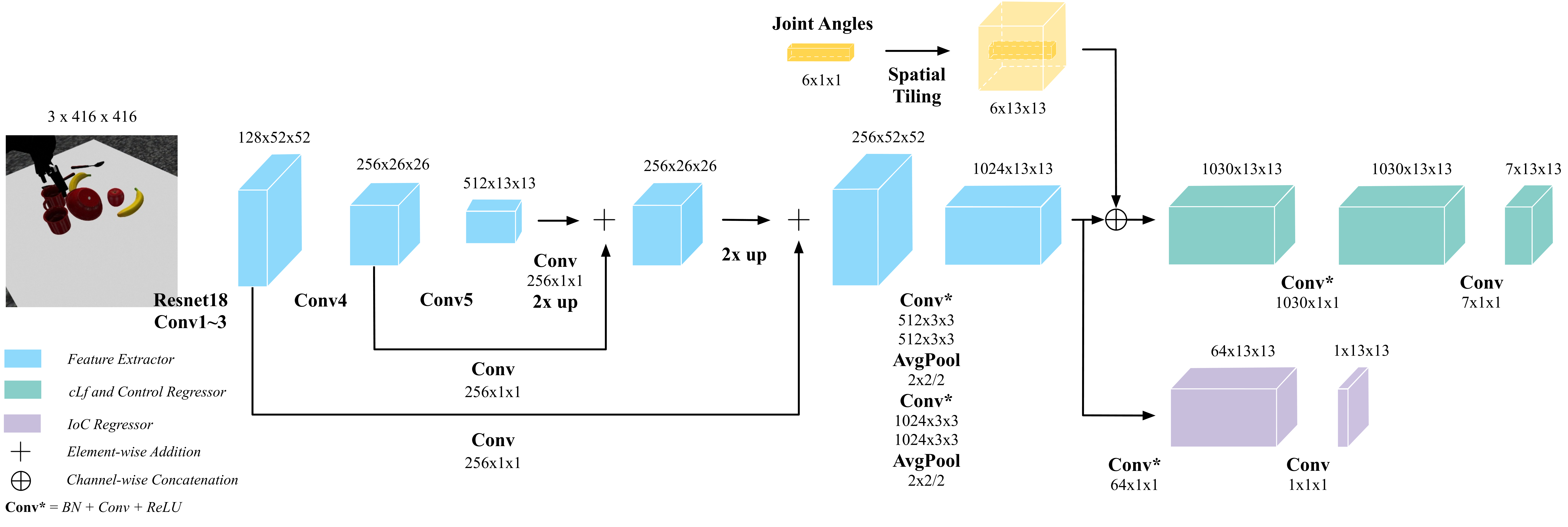}
		\label{fig:net_arch_main}}\\
	\subfloat[]{
		\includegraphics[width=0.2\textwidth]{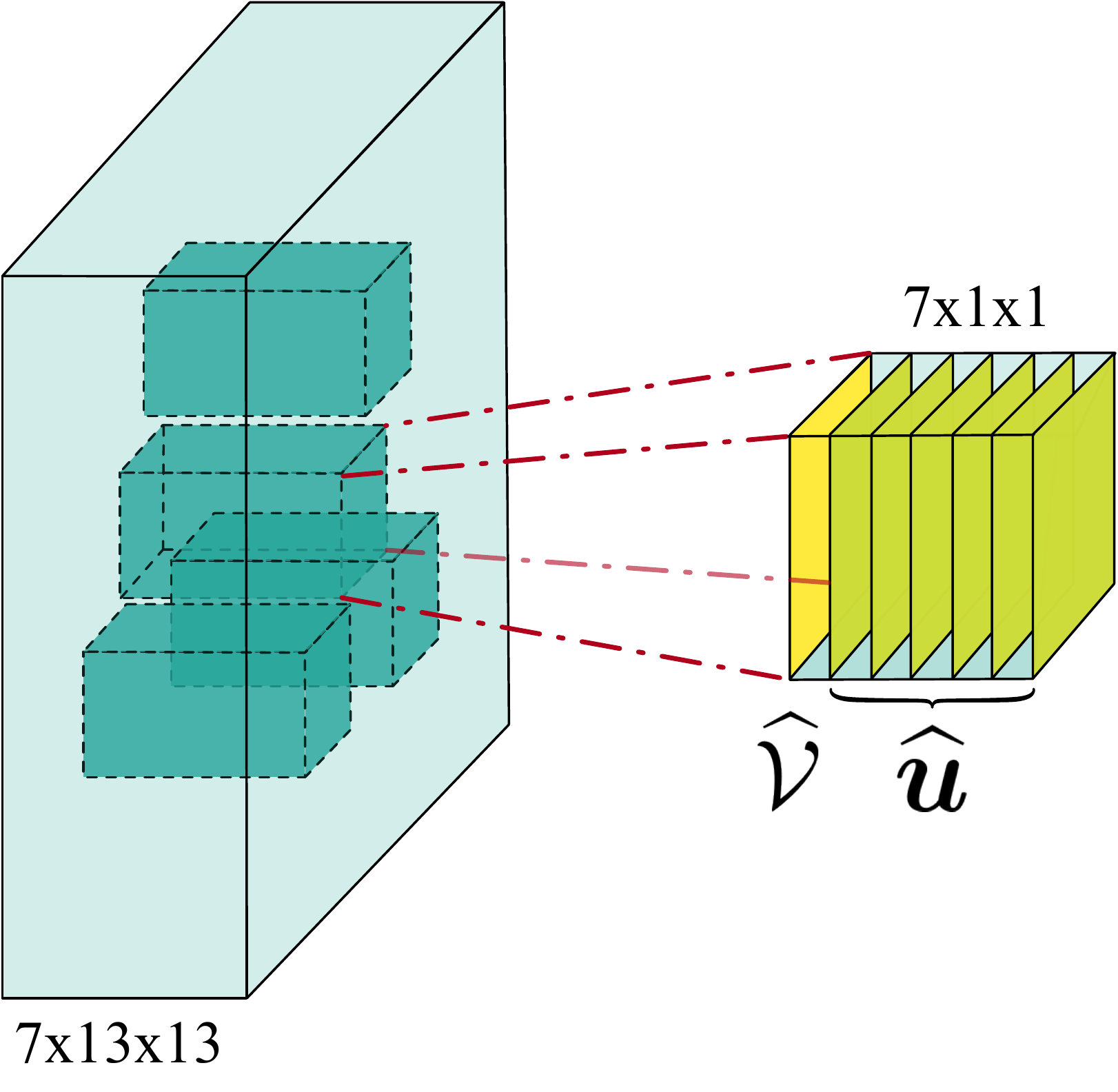}
		\label{fig:net_arch:clf_regressor}}
	\hspace{3mm}
	\subfloat[]{
		\includegraphics[width=0.18\textwidth]{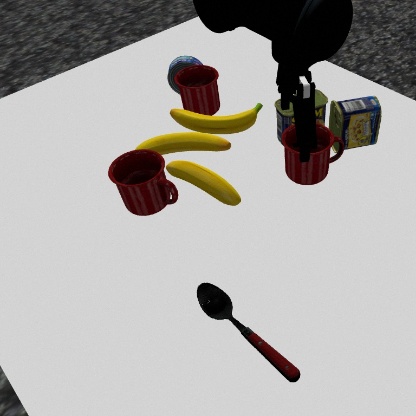}
		\label{fig:net_arch:sim_sample}}
	\hspace{3mm}
	\subfloat[]{
		\includegraphics[width=0.18\textwidth]{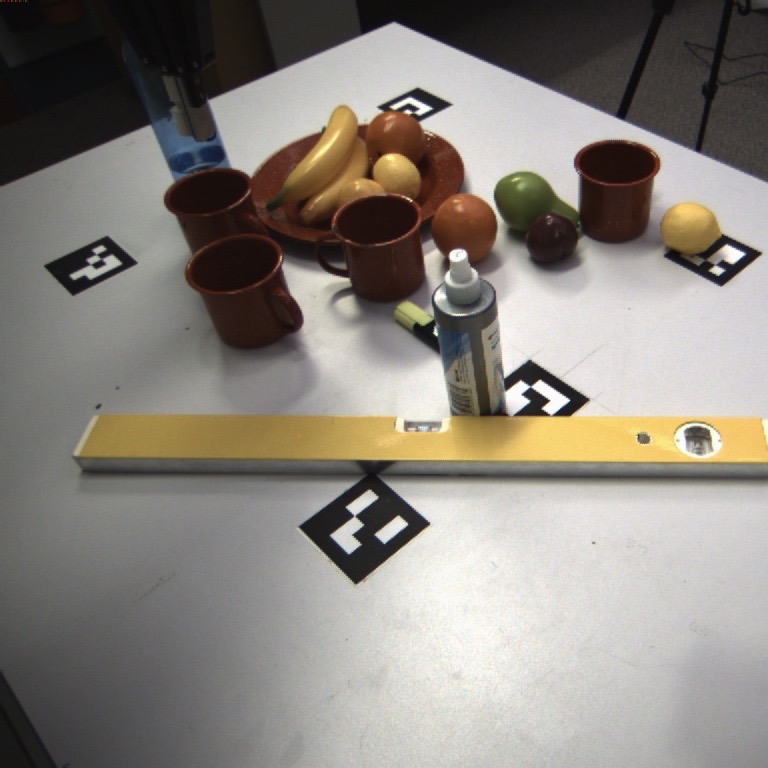}
		\label{fig:net_arch:input}}
	\hspace{3mm}
	\subfloat[]{
		\includegraphics[width=0.18\textwidth]{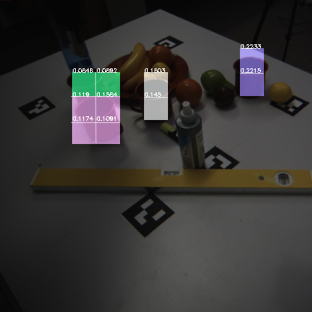}
		\label{fig:net_arch:net_output}}
	\caption{
		(a) The proposed network architecture. (b) The output tensor structure of the cLf regressor contains the value of cLf $\widehat{\lyap}$ and velocity control $\widehat{\bm{u}}\in \R^6 $ in each cell. (c) An example of the simulated training image with four targets (mugs). (d) An example of the real test image with four targets. (e) Visualisation of the IoC regressor output given the input image shown in Fig.\ref{fig:net_arch:input} and corresponding joint angles; cells with IoC scores higher than 0.6 are colour-coded based on clustering results from affinity propagation~\cite{frey2007clustering} given the corresponding vector $(\widehat{\lyap},\ \widehat{\bm{u}}^{\top})\in\R^7$.}
	\label{fig:apcNet}
	\vspace{-2mm}
\end{figure*}
\subsection{Velocity Controller Design}
To formulate the velocity control, we use the velocity Jacobian $\bm{J} = \bm{J}(\bm{\theta})$ for the manipulator.
Denoting the angular and translational rigid body velocity of $\pose{H}{}$ expressed in its body-fixed frame $\{H\}$ by $\bm{\omega}\in \R^{3\times1}$ and $\bm{v}\in\R^{3\times1}$ respectively, one has $(\bm{\omega}, \bm{v})^\top = \bm{J} \dot{ \bm{\theta}}$.

Let
\[
\begin{pmatrix}
\bm{\omega}_{\times}& \bm{v} \\
\bm{0}_{1\times3} & 0 \\
\end{pmatrix}^\vee = (\bm{\omega}, \bm{v})^\top,
\]
where $[\ ]_{\times}$ denotes the skew-symmetric matrix of a vector.  The partial differential of cLf $\lyap$~\eqref{eq:cLf_complex} with respect to the pose of the world frame relative to end-effector $\pose{}{H}$ frame is derived from Eq.~\eqref{eq:cLf_complex} as
\begin{equation*}
        \nabla{\lyap{}} = \text{proj}_{\se(3)}\left(\pose{H}{}^\top(\pose{H}{}-\pose{G}{}^*)\right) \in \se3,
\end{equation*}
where $\nabla$ is differentiation with respect to $\SE(3)$ and $\text{proj}_{\se(3)}$ is the matrix projection operator that maps an arbitrary $4\times4$ matrix to the nearest member of $\se(3)$ measured in Frobenius norm.
The proposed joint velocity control is
\begin{equation}
    \label{eq:vel_ctrl}
    \bm{u} := -\bm{J}^{-1}(\nabla{\lyap{}})^\vee.
\end{equation}
With this, one guarantees
\begin{equation*}
    \dot{\lyap}  = -\text{tr}(\nabla{\lyap{}}^\top\nabla{\lyap{}})
    = -\|\nabla{\lyap{}}\|^2_F<0.
\end{equation*}

In addition to using the cLf for control design, we also use the first order differential structure of the cLf as a regularisation term later in the paper.
Consider a small arbitrary variation $\delta\bm{\theta}$ in the joint coordinates $\bm{\theta}$.
The first order variation in the Lyapunov function can be written
\begin{equation*}
\frac{\partial \lyap}{\partial\bm{\theta}}^\top \delta\bm{\theta}
= \text{tr}(\nabla\lyap^\top(\bm{J}\delta\bm{\theta})^\wedge)
= -\text{tr}\left(((\bm{J}\bm{u})^\wedge)^\top(\bm{J}\delta\bm{\theta})^\wedge\right),
\end{equation*}
where $(\cdot)^\wedge$ denotes the mapping from $\R^{6\times1}$ to $\se(3)$ opposed to $(\cdot)^\vee$ defined earlier and $u$ is given by \eqref{eq:vel_ctrl}.
From the first-order Taylors expansion, one has
\begin{align}
\lyap(\bm{\theta} + \delta\bm{\theta}) - \lyap(\bm{\theta})
& \approx - \text{tr}\left(((\bm{J}\bm{u})^\wedge)^\top(\bm{J}\delta\bm{\theta})^\wedge\right) \notag \\
& = -  \delta\bm{\theta}^\top \bm{J}^\top\bm{Q}\bm{J}\bm{u},
\label{eq:differential_constraint}
\end{align}
where the diagonal matrix 
\[
\bm{Q} =  \begin{pmatrix}
    2\cdot\eye_3 & \bm{0}_{3\times 3} \\
    \bm{0}_{3\times 3} & \eye_3 \\
    \end{pmatrix},
\]



\section{Learning the Control Lyapunov Function}\label{sec:III:learning}

In this section, we describe the network architecture (Fig.\ref{fig:net_arch_main}) and the learning strategy to predict the control Lyapunov function and corresponding velocity control that acts as inputs for the closed-loop system. Our network architecture is inspired by YOLO \cite{redmon2016you, redmon2017yolo9000} since single-shot object detectors achieve impressive real-time performance.
Moreover, we introduce a Siamese regression network to enforce the differential constraint in Eq.~\eqref{eq:differential_constraint} in training.

\subsection{Network Architecture}
\textbf{The feature extractor} is based on ResNet18 \cite{He:2016tt}. Rather than only using its last convolutional features, we combine the output features from last three residual blocks, namely conv3, conv4, conv5, into 256-channel multiscale feature maps by adapting the top-down architecture with lateral connections (feature pyramid network) proposed by~\cite{lin2017feature}.
Another $3\times3$ convolutional layer is employed afterwards to fuse the multiscale image features.
The \textbf{joint angles} $\bm{\theta}\in\R^{6\times1}$ are tiled over the spatial dimension and concatenated with the extracted image features along the channel dimension. \textbf{The regressor} comprises two branches: one branch scores object presence, the other regresses the cLf value $\widehat{\lyap}$ and velocity control $\widehat{\bm{u}}$. In this way, our proposed network is able to predict not only confidence of target presence but also control inputs.

As shown in Fig.\ref{fig:net_arch_main}, our fully-convolutional network divides an input image into $13\times 13$ grids, each grid cell predicts a target presence score, a cLf value and velocity control. A zoomed-in output structure of the control regressor is depicted in~Fig.\ref{fig:net_arch:clf_regressor}.  Note that, all the cells belong to the same object should generate identical cLf value and control.

\subsection{Loss Function Formulation}
Since we do not predict bounding boxes in our method, the intersection over union (IoU) metric is not applicable. Instead, we define the confidence metric as intersection over cell (IoC): the ratio of the intersection between a 2D ground-truth bounding box and the occupied cell girds, denoted as $\mathcal{C}$.
\begin{figure}[t]
	\centering
	\includegraphics[width=0.49\textwidth]{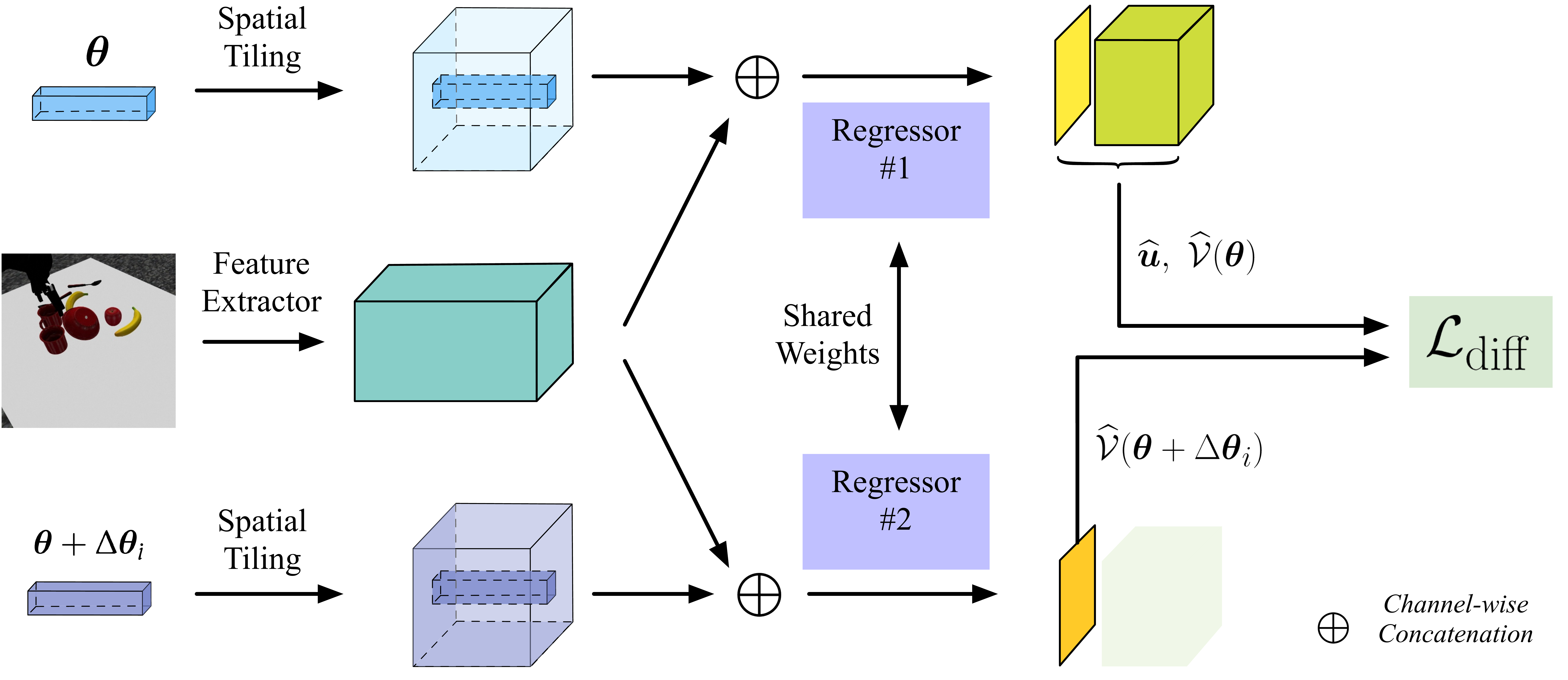}
	\caption{Siamese regression for computing the $i^{\text{th}}$ term of $\C{L}_{\text{diff}}$ in Eq.~\eqref{eq:loss_diff} given an input image and corresponding joint angle $\bm{\theta}$. Two identical regressors share the same weights and the extracted image feature. cLf and control inferred from Regressor \#1 and the corresponding cLf computed from Regressor \#2 given a perturbation $\Delta\bm{\theta}_i$  then become the inputs of the differential loss term.}
	\label{fig:Siamese}
	\vspace{-3mm}
\end{figure}

Our loss function consists of a multi-task regression loss $\mathcal{L}_{\text{reg}}$ in Eq.~\eqref{eq:loss_reg} and a differential constraint regularisation term $\C{L}_{\text{diff}}$ in Eq.~\eqref{eq:loss_diff}. Similar to YOLO, $\mathcal{L}_{\text{reg}}$ is a sum of multiple mean absolute error terms:
\begin{multline}
    \label{eq:loss_reg}
    \C{L}_{\text{reg}} = \frac{1}{\sum_{k=1}^{S^2}\mathbb{1}^{\text{obj}}_k} \sum_{k=1}^{S^2}\mathbb{1}^{\text{obj}}_k \left(\lambda_{\lyap}\left|\lyap_k-\widehat{\lyap}_k\right|\right.+\\
    \frac{\lambda_{\bm{u}}}{6}\left|\bm{u}_k-\widehat{\bm{u}}_k\right|+
    \left.\lambda_{\text{obj}}\left|\C{C}_k-\widehat{\C{C}}_k\right|\right)\\+ \frac{\lambda_{\text{no\_obj}}}{\sum_{k=1}^{S^2}\mathbb{1}^{\text{no\_obj}}_k}
    \sum_{k=1}^{S^2}\mathbb{1}^{\text{no\_obj}}_k\left|\C{C}_k-\widehat{\C{C}}_k\right|,
\end{multline}
where an indicator function $\mathbb{1}^{\text{obj}}$ denotes a target presents in cell $k$ , and $\mathbb{1}^{\text{no\_obj}}$ is its binary complement. $S^2$ denotes the number of cells in the final prediction.

In the context of manipulation, all the joints contribute to the end-effector pose via manipulator's kinematic chain.
This correlation among the control variables also \textbf{needs to be learnt} by designing a suitable loss function.
The control Lyapunov function inherently correlates the control input with the cLf value.
To capture this correlation we propose an additional regularisation term $\mathcal{L}_{\text{diff}}$ that \textbf{explicitly encourages} the differential constraint formulated in Eq.~\eqref{eq:differential_constraint}.
Let $\bm{e}_i,\text{ for } i = 1, \ldots, 6$, denote the $i^{\text{th}}$ unit vector of standard basis in $\R^6$ and $\delta\bm{\theta}_i = \delta\theta_i \bm{e}_i$ represents the joint perturbation at $i^{th}$ joint.
Define
\begin{multline}
     \label{eq:loss_diff}
     \C{L}_{\text{diff}} := \frac{\lambda_{\text{diff}}}{6\sum_{k=1}^{S^2}\mathbb{1}^{\text{obj}}_k}
     \sum_{k=1}^{S^2}
     \sum_{i=1}^{6}
     \mathbb{1}^{\text{obj}}_k
     \bigg|\bm{e}_i^\top\bm{J}^\top\bm{Q}\bm{J}\widehat{\bm{u}}\ +\\
     \delta\theta^{-1}_i\big(\widehat{\lyap}({\bm{\theta} + \delta \theta}_i\bm{e}_i )-\widehat{\lyap}(\bm{\theta})\big)\bigg|.
\end{multline}
where we separately apply Eq.~\eqref{eq:differential_constraint} in each of the joint directions and scale by  $(\delta \theta_i)^{-1}$ to make $\mathcal{L}_{\text{diff}}$ comparable to $\mathcal{L}_{\text{reg}}$.

Assuming that the change in observed image is negligible when there is a small perturbation in the robot joint angles, we propose a \textbf{Siamese regressor}, as shown in Fig.\ref{fig:Siamese}.
In the regressor network, the same image input is shared by both perturbed and unperturbed branches.
In each branch, the joint angle vector is tiled along the spatial dimension and then concatenated with the image features along the feature channel dimension.
The Jacobian $\bm{J}$ in Eq.\eqref{eq:loss_diff} is computed analytically by the forward kinematic and each branch predicts cLf and velocity control from its concatenated features.
The outputs of these two branches are fed into the differential constraint in Eq.~\eqref{eq:loss_diff}.
Note that, the Siamese regressor is not used at test time.

Combining the multi-task regression loss function in Eq.\eqref{eq:loss_reg} and the regularisation term in Eq.\eqref{eq:loss_diff}, we obtain the total objective
\begin{equation}
\label{eq:loss}
	\C{L} = \C{L}_{\text{reg}} + \C{L}_{\text{diff}}.
\end{equation}


%
%


\section{Implementation}
\label{sec:implementation}
\begin{figure}[t]
	\centering
	\subfloat[Simulator Set-up]{
		\includegraphics[width=0.18\textwidth]{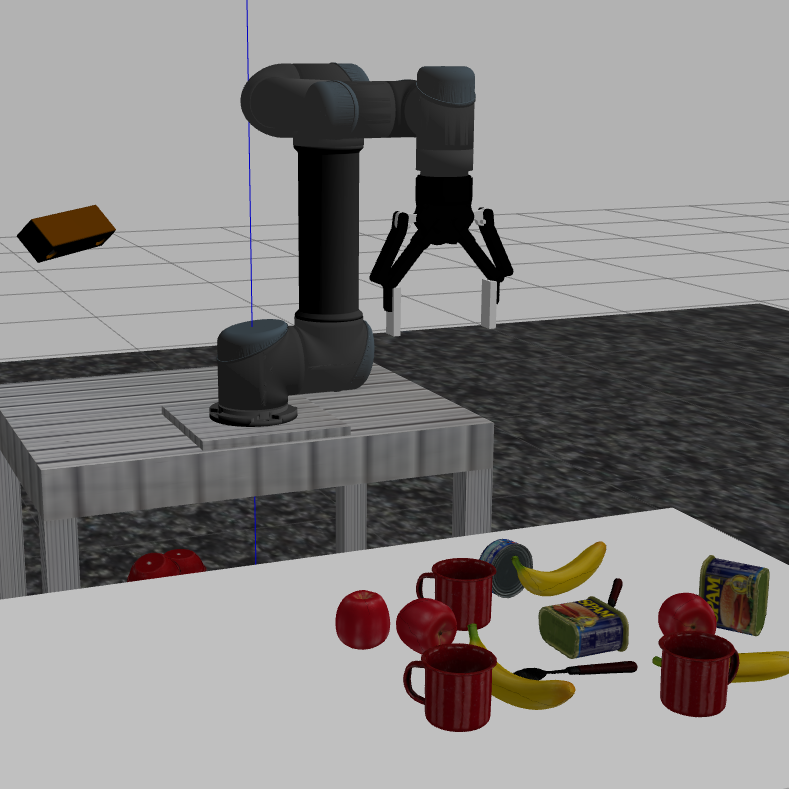}
		\label{fig:lab_setup:sim}}
	\hspace{3mm}
	\subfloat[Real-world Set-up]{
		\includegraphics[width=0.18\textwidth]{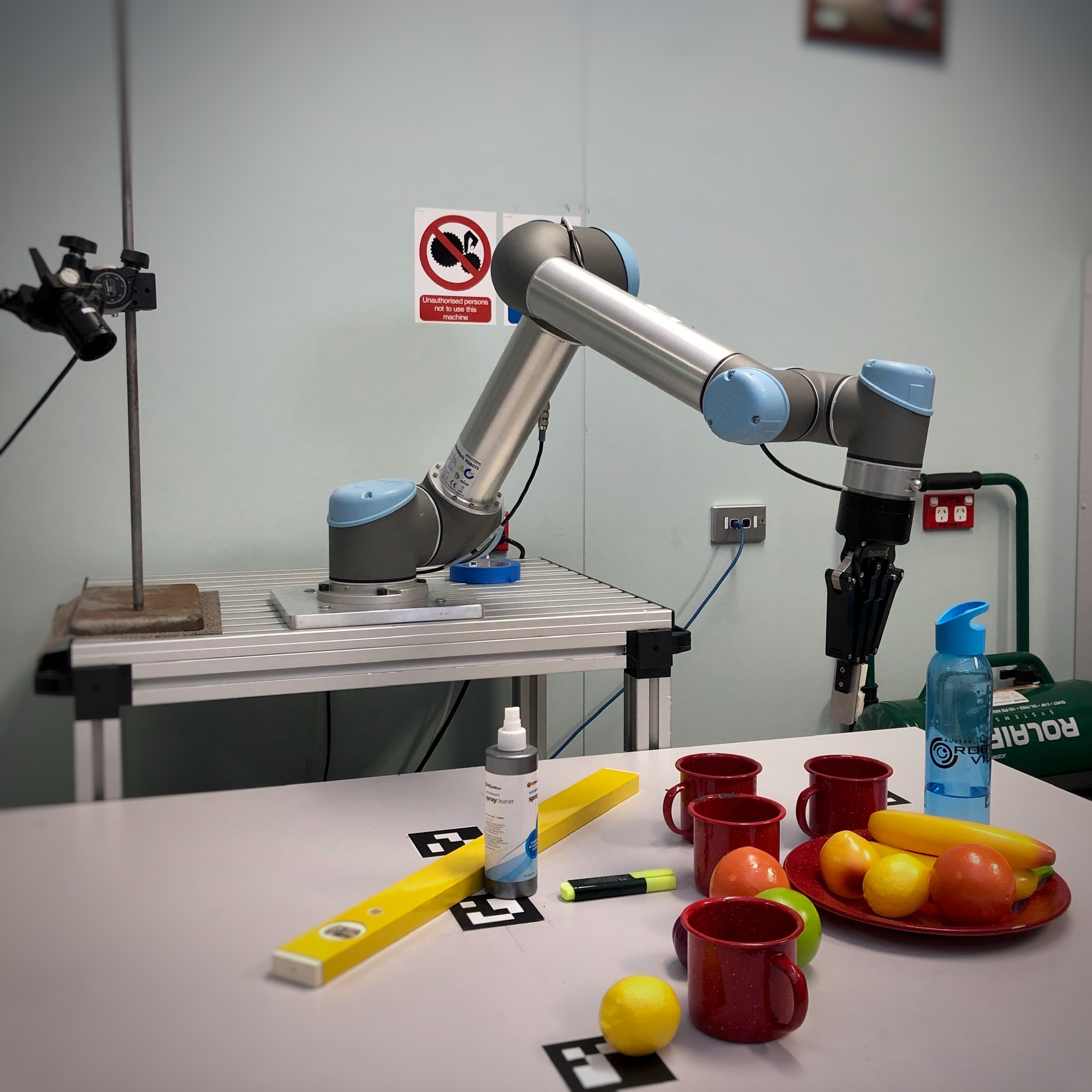}
		\label{fig:lab_setup:real}}
	\caption{Lab environment set-up: the first-person camera is positioned as shown, pointing towards the table workspace. Examples of captured images are shown in Fig.\ref{fig:lab_setup:sim} and \ref{fig:lab_setup:real}. The simulated environment is geometrically identical (within measurement error) to the lab. The simulated camera is calibrated to the real counterpart.}
	\label{fig:lab_setup}
	\vspace{-3mm}
\end{figure}

To perform the reaching and grasping experiments, we use a UR5 six DoF manipulator with a two-finger parallel gripper. In order to increase the contact area and surface friction, customised finger tips with textured soft-silicone pads are attached to the gripper. The vision sensor is a 50Hz RGB camera of $1280\times768$ pixel resolution, which constrains the closed-loop control to 50Hz.
Since the entire workspace only occupies a region of $768\times768$ pixels, we crop the region from the original images and then resize it to the required input size of our network.

\subsection{Simulator}
\label{subsec:simulator}
Since collecting real-world manipulator data is a costly process \cite{Levine:2017cu},
trainining a network with large-scale simulated data is a viable alternative to collecting a large amount of real world data \cite{Fangyi2018,Viereck:2017uq,peng2017sim}. We replicate the real-world setting in Gazebo simulator~\cite{gazebo2004}, as it provides consistent control interface between the simulator and the real hardware (see Fig.\ref{fig:lab_setup}).
	
We jointly calibrate the camera intrinsics, camera pose and tabletop height with respect to the manipulator base frame via an automatic procedure.
A calibration board is attached to the end-effector. The joint angles and the images of the calibration board are captured for calibrating the camera pose. We also place the board on the tabletop to calibrate the tabletop height.
The camera poses and tabletop height are optimised with collected images. \zheyu{\small{Code of the calibration package is available at: \url{https://github.com/Zheyu-Zhuang/ur5_joint_calib_toolkit}}}

\subsection{Data Collection}
\label{subsec:dset}
In this work, we use up to three identical targets to generate the multi-instance scenarios.
Sampling the location of the manipulator in the training set plays an important role to improve the accuracy of the resulting network as well as the effeciency of the training procedure.
Thus, we sample the manipulator locations coarsely to improve the training efficiency when the manipulator is far away from goals, while sampling more densely in the neighbouring regions of the goal to improve the final positioning of the end-effector.

Each end-effector translation component is sampled from a zero-mean uniform distribution bounded by the workspace.
At each end-effector position, the Euler angles of the end-effector frame are sampled individually on uniform distributions and constrained so that the end-effector always points towards the workspace.

Given a random end-effector pose, we randomly select one of the instances and execute the proposed controller for a duration sampled from a uniform distribution. One image and its corresponding cLf value and velocity control, as one sample, are collected at the end of the reaching trajectory.
\begin{table}
	\begin{center}
		\begin{tabular}{c c c c c c}
			\toprule
			\textbf{\makecell{}} &\multicolumn{2}{c}{\textbf{\makecell{Mean Abs. \\ Error}}} & \multicolumn{2}{c}{\textbf{\makecell{Mean Relative \\ Error}}}& \textbf{\makecell{Diff.\\ Error}}\\
			& \begin{small}$\widehat{\lyap}(\bm{\theta})$\end{small} & \begin{small}$\widehat{\bm{u}}$ \end{small} & \begin{small}$\widehat{\lyap}(\bm{\theta})$ \end{small} &  \begin{small}$\widehat{\bm{u}}$\end{small} & \\
			\midrule
			\textbf{w} $\mathcal{L}_{\text{diff}}$ & $\bm{0.017}$ & $\bm{0.035}$ & $\bm{0.404}$ & $\bm{0.281}$ & $\bm{0.098}$ \\
			\textbf{w/o} $\mathcal{L}_{\text{diff}}$& 0.020 & 0.036 &  0.441 & 0.301 & 0.113 \\
			\bottomrule
		\end{tabular}
	\end{center}
	\vspace{-3mm}

	\caption{Comparison of networks trained with/without the proposed differential constraint. Mean differential error is defined in Eq~\eqref{eq:loss_diff}.}
	\label{tab:nn_results}
	\vspace{-3mm}
\end{table}
\subsection{Learning Details and Results}
\label{subsec:learn_results}
Our training dataset contains 17K simulated scenes, with on average 5 samples per scene.
We train our networks with 90\% of the training dataset while using the remaining 10\% for evaluation.
Random cropping, rotation, and colour jittering are employed during training.
Specifically, we randomly resize an input image with a scaling factor between 0.95 and 1.05 of its original resolution.
After rescaling, we crop or pad the image to its original size.
Random rotations in the range $-1.5\degree$ and $1.5\degree$ are applied afterwards to make the algorithm tolerant to real-world hardware calibration errors.
The brightness, saturation, contrast and hue of input images are randomly jittered at 10\% of their maximum ranges to alleviate the domain gap between simulated and real data.
The ResNet18~\cite{He:2016tt} backbone in Fig.~\ref{fig:net_arch_main} is pre-trained on ImageNet \cite{ImageNet}. We use ADAM~\cite{kingma2014adam} optimiser with the batch size 64. The learning rate is initialised as $10^{-3}$ with a decay rate 0.1 for every 15 epochs.

Numerical differentiation achieves more accurate approximation to analytic gradients when the perturbation is small.
However, small perturbations significantly affect the numerical stability of numerical gradients in the differential learning cost, especially during the early-stage training.
To address this, we initialise the joint perturbation $\delta\theta$ with 0.05 rad, and reduce it by 0.002 rad after each epoch until it reaches the minimum value $\delta\theta = 0.003$.

In training our proposed model, the hyper-parameters, $\lambda_{\lyap},\ \lambda_{\bm{u}},\ \lambda_{\text{obj}}\,\ \lambda_{\text{no\_obj}} \ \text{and}\  \lambda_{\text{diff}}$, are empirically set to 0.2, 1, 0.8, 0.08 and 0.1 respectively. To evaluate the effect of the differential constraint~Eq.\eqref{eq:loss_diff}, we train another network, marked as vanilla regression network, without exploiting the constraint. To be specific, we set the weight $\lambda_{\text{diff}}$ to 0.

The evaluation results are shown in Table \ref{tab:nn_results}. The Mean Absolute Error is employed to measure the difference between the estimated and ground-truth control variables, and provides an overall indication of the quality of the approximated control.
The Mean Relative Error (MRE) is
\[
\text{MRE} = \frac{ | y - \widehat{y}|}
{|y|+\epsilon},
\]
where $\epsilon=10^{-3}$ is a small positive constant to avoid division by zero and $y$ and $\widehat{y}$ represent the groud-truth and estimation respectively. MRE is sensitive to the changes of variables but insenstive to their magnitudes. Thus, MRE is a better measurement for velocity control. 

The results in Table \ref{tab:nn_results} demonstrate that exploiting our proposed differential constraint produces lower errors for all metrics.
Since our method is able to establish the first-order derivative relationship among control variables explicitly by using our regularisation term, our network captures the structure of the control, and its underlying link to the differential of a Lyapunov function, more effectively than the vanilla regression network.
The advantage of applying the differential constraint is further evaluated in real-world reaching experiments in Sec.\ref{subsec:diff_eval_real}.

\section{Experiments and Results}
\label{sec:results}
Here, we demonstrate the efficacy of the proposed
method on real-world reaching/grasping by comparing with a state-of-the-art detection-based RGB pose estimation algorithm and a vanilla regression baseline. The grasping strategy for mugs is designed as opening the gripper when its tip is inside the cavity of mugs.
\subsection{Pose-estimation Baseline}
\label{sec:siam_vs_dir}
We implement the real-time single shot 6D-pose network proposed by Tekin \textit{et. al}~\cite{Tekin:2017wk}, a state-of-the-art RGB pose-estimation algorithm, and estimate its upper-bound grasping rate in static environment by assuming the system has a perfect pose-based controller and knows the grasping height.

For a fair comparison, this baseline is retrained only with the synthetic data collected from the simulator as described in Sec.\ref{subsec:learn_results}. The dataset contains 29K training images and 1K test images of one mug randomly positioned on the tabletop in cluttered scenes. We employ the same training protocols as in \cite{Tekin:2017wk} to re-train the object pose estimation network.

Due to the rotational symmetric property of mugs, we use ADD-S~\cite{xiang2017posecnn} to evaluate the pose estimation accuracy of the mug. ADD-S measures the mean closest point distance between the 3D model vertices transformed by the ground-truth pose and their counterparts transformed by the estimated pose. To be specific, ADD-S is defined as follow:
\begin{equation*}
\textbf{ADD-S} = \frac{1}{m}\sum_{x_1\in\mathcal{M}}\min_{x_2\in\mathcal{M}}\|\bm{R}\bm{x_1}+\bm{T} - (\widehat{\bm{R}}\bm{x_2}+\widehat{\bm{T}})\|,
\end{equation*}
where $m$ denotes the number of 3D object points and $\mathcal{M}$ represents the 3D model vertices. \zheyu{$\bm{R}\in\SO(3)$ and $\bm{T}\in\R^3$ denotes the rotation and translation respectively.}
The estimation is considered as correct if the ADD-S score is less than 10\% of the diameter of the mug (\ie12mm), a level that would correspond to a successful grasp in the real-world.
The network~\cite{Tekin:2017wk} achieves 91.5\% success on the evaluation dataset.

The real-world evaluation for the baseline pose estimation was considerably poorer than in simulation.
Table~\ref{tab:baseline_results} shows a 30mm pose error success rate for the case where there are one, two, three or four instances of mugs in the workspace.
The poor performance of the pose estimation in the real world data is related to high levels of false positives that can be traced to poor generalisation error from the training sequence to the data.
The increase in pose estimation accuracy with the number of instances in the scene is associated with the decrease in false positives, simply because more of the scene is occupied by true instances.
This could be improved by transfer learning from real world data, however, the proposed algorithm does not require transfer learning and the comparison provided is fair.
The performance of pose estimation in the real-world scenario was insufficient to implement a reliable grasping control and we will compare with the provided estimated success rates as an upper bound on the best real-world results that could be expected from this approach.


\begin{table}
	\begin{center}
		\begin{tabular}{c c c c c c c}
			\toprule
			& \multicolumn{5}{c}{\textbf{2D translation $\bm{<}$ 3 cm (\%)}}  \\		
			& \small{1} & \small{2} & \small{3} & \small{4} & \small{average} \\			\midrule
			& 17.5 & 27.5 &  45.0 & 45.0 & 33.75\\
			\bottomrule
		\end{tabular}
	\end{center}
	\vspace{-1mm}
	\caption{ Evaluation of success rate for single shot pose estimation baseline~\cite{tekin2018real} trained on the synthetic dataset.
The network is tested on 180 scenes in the real lab environment, for one, two, three, and four mugs shown as separate columns.}
	\label{tab:baseline_results}
	\vspace{-4mm}
\end{table}
\subsection{Robustness against false positives detections}

In our proposed algorithm, the vision pipeline only estimates a coarse object presence score that helps to separate the target class from the clutter.
The estimation of cLf carries out the heavy lifting: selecting the control corresponds to the minimum cLf and establishing the relationship among the control variables explicitly with our proposed differential constraint.
In our experiments, the cLf of most false positives in the vision pipeline larger than the current minimum will be rejected by the system.

Cases of false positives exhibiting lower values of cLf are significantly suppressed with help of the closed-loop controller design.
Inspired by momentum terms widely used in the optimisation algorithms, we implement the controller $\bar{\bm{u}}$ as:
\begin{equation*}
\bar{\bm{u}}_t = \eta\bar{\bm{u}}_{t-1} + (1-\eta)\widehat{\bm{u}}_t,
\end{equation*}
where $\eta$ is a tuneable constant $\in[0, 1]$ and $\widehat{\bm{u}}_t$ is a current raw velocity control prediction from the network. The constant $\eta$ is set to 0.6 in our experiments. It leads to smother reaching trajectories especially when increasing the controller gain. 
More importantly, the momentum in the velocity controller keeps the end-effector moving towards the target for a short duration even in the presence of false positives.

\begin{table}[t]
	\begin{center}
		\begin{tabular}{c c c c c c c}
			\toprule
			\textbf{\makecell{\# \\ Objs}}& $\bm{\mathcal{L}_{\text{diff}}}$ & \multicolumn{2}{c}{\textbf{\makecell{Mean Abs. \\ Error}}} & \multicolumn{2}{c}{\textbf{\makecell{Mean Relative \\ Error}}}& \textbf{\makecell{GSS \\ (\%)}}\\
			
			&  & \begin{small}$\widehat{\lyap}(\bm{\theta})$\end{small} & \begin{small}$\widehat{\bm{u}}$ \end{small} & \begin{small}$\widehat{\lyap}(\bm{\theta})$ \end{small} &\begin{small}$\widehat{\bm{u}}$\end{small} & \\
			\midrule
			\multirow{2}{*}{\textbf{1}}
			&\textbf{w} & $\bm{0.01}$ & $\bm{0.06}$ & $\bm{0.78}$& $\bm{0.89}$ &  $\bm{94.0}$ \\
			& \textbf{w/o} & $0.02$ & $0.07$ & $1.99$ & $1.14$ & 79.0 \\
			\midrule
			\multirow{2}{*}{\textbf{2}}
			&\textbf{w} & $\bm{0.01}$ & $\bm{0.08}$ & $\bm{0.83}$& $\bm{1.24}$ &  $\bm{91.0}$ \\
			& \textbf{w/o} & $0.01$ & $0.12$ & $0.98$ & $1.39$ & 73.0  \\
			\midrule
			\multirow{2}{*}{\textbf{3}}
			&\textbf{w} & $\bm{0.01}$ & $\bm{0.09}$ & $\bm{0.89}$& 		$\bm{1.12}$ &  $\bm{87.0}$ \\
			& \textbf{w/o} & $0.01$ & $0.10$ & $1.21$ & $1.48$ & 78.0  \\
			\midrule
			\multirow{2}{*}{\textbf{4}}
			&\textbf{w} & $0.01$ & $\bm{0.09}$ & $\bm{0.72}$ & $\bm{1.17}$ &  $\bm{89.0}$ \\
			& \textbf{w/o} & $0.01$ & $0.10$ & $0.80$ & $1.39$ & 78.0  \\
			\midrule
			\multirow{2}{*}{\textbf{All}}
			
			&\textbf{w} & $\bm{0.01}$ & $\bm{0.08}$ & $\bm{0.80}$ & $\bm{1.10}$ &  $\bm{90.3}$ \\
			& \textbf{w/o} & $0.01$ & $0.10$ & $1.24$ & $1.35$ & 77.0  \\
			\bottomrule
		\end{tabular}
	\end{center}
	\vspace{-1mm}
	\caption{Statistical Results of 40 grasping experiments (10 samples for one, two, three and four instances) in real-world using with/without the proposed differential constraint. Each experiment has 10 reaching trajectories starting from a random initial position, only one grasping attempt for each trajectory. }
	\label{tab:nn_real_world}
	\vspace{-4mm}
\end{table}

\subsection{Effectiveness of the Differential Constraint}
\label{subsec:diff_eval_real}
The experiment has 40 scenes -- ten for one, two, three, and four targets in cluttered scenes. For a given scene, we test both our method and vanilla regression counterpart, and the end-effector is initialised at a random position using the strategy described in Sec.~\ref{subsec:dset} 10 times. We record 50 Hz closed-loop reaching trajectories at frequency of 25Hz. This yields total 400 reaching trajectories for each method.

The results in Table~\ref{tab:nn_real_world} indicates our proposed method achieves 90.3\% grasp success rate without leveraging any additional simulation-to-real transfer techniques.
This is compared to the estimated 33.75\% success rate of the pose estimation algorithm documented in Table \ref{tab:baseline_results}.
Moreover, different from the pose estimation based method, the grasp success rate of our proposed algorithm remains consistent as the number of target instances varies.
This demonstrates the superior multi-instance capability of our method.

\zheyu{Incorporating the differential constraint improves the grasp success rate by $13.3\%$. In addition, the mean relative errors associate with both cLf and control are also reduced by large margins. This demonstrates the effectiveness of the proposed differential constraint.}

\section{Conclusion}
We proposed a closed-loop, multi-instance control algorithm for visually guided reaching. The system is able to reach and grasp with 90.3\% accuracy in cluttered environments from a over-the-shoulder RGB camera up-to 85Hz. The inclusion of a first-order differential constraint associated with the control Lyapunov function leads to better reaching/grasping performance in comparison to vanilla regression and pose estimation based baselines. Our system is robust against false positive detections caused by the sim-to-real domain gap.



\bibliographystyle{IEEEtran}
\bibliography{IEEEabrv,zheyu_icra_2020}

\begin{thebibliography}{10}
\providecommand{\url}[1]{#1}
\csname url@rmstyle\endcsname
\providecommand{\newblock}{\relax}
\providecommand{\bibinfo}[2]{#2}
\providecommand\BIBentrySTDinterwordspacing{\spaceskip=0pt\relax}
\providecommand\BIBentryALTinterwordstretchfactor{4}
\providecommand\BIBentryALTinterwordspacing{\spaceskip=\fontdimen2\font plus
\BIBentryALTinterwordstretchfactor\fontdimen3\font minus
  \fontdimen4\font\relax}
\providecommand\BIBforeignlanguage[2]{{%
\expandafter\ifx\csname l@#1\endcsname\relax
\typeout{** WARNING: IEEEtran.bst: No hyphenation pattern has been}%
\typeout{** loaded for the language `#1'. Using the pattern for}%
\typeout{** the default language instead.}%
\else
\language=\csname l@#1\endcsname
\fi
#2}}

\bibitem{kemp2007challenges}
C.~C. Kemp, A.~Edsinger, and E.~Torres-Jara, ``Challenges for robot
  manipulation in human environments [grand challenges of robotics],''
  \emph{IEEE Robotics \& Automation Magazine}, vol.~14, no.~1, pp. 20--29,
  2007.

\bibitem{hutchinson1996tutorial}
S.~Hutchinson, G.~D. Hager, and P.~I. Corke, ``A tutorial on visual servo
  control,'' \emph{IEEE transactions on robotics and automation}, vol.~12,
  no.~5, pp. 651--670, 1996.

\bibitem{corke1993visual}
P.~I. Corke, ``Visual control of robot manipulators--a review,'' in
  \emph{Visual Servoing: Real-Time Control of Robot Manipulators Based on
  Visual Sensory Feedback}.\hskip 1em plus 0.5em minus 0.4em\relax World
  Scientific, 1993, pp. 1--31.

\bibitem{corke2000real}
P.~I. Corke and S.~A. Hutchinson, ``Real-time vision, tracking and control,''
  in \emph{Robotics and Automation, 2000. Proceedings. ICRA'00. IEEE
  International Conference on}, vol.~1.\hskip 1em plus 0.5em minus 0.4em\relax
  IEEE, 2000, pp. 622--629.

\bibitem{Rad:2017et}
M.~Rad and V.~Lepetit, ``{BB8: A Scalable, Accurate, Robust to Partial
  Occlusion Method for Predicting the 3D Poses of Challenging Objects without
  Using Depth},'' in \emph{2017 IEEE International Conference on Computer
  Vision (ICCV)}.\hskip 1em plus 0.5em minus 0.4em\relax IEEE, 2017, pp.
  3848--3856.

\bibitem{Tekin:2017wk}
B.~Tekin, S.~N. Sinha, and P.~Fua, ``{Real-Time Seamless Single Shot 6D Object
  Pose Prediction},'' \emph{arXiv.org}, Nov. 2017.

\bibitem{brachmann2014learning}
E.~Brachmann, A.~Krull, F.~Michel, S.~Gumhold, J.~Shotton, and C.~Rother,
  ``Learning 6d object pose estimation using 3d object coordinates,'' in
  \emph{European conference on computer vision}.\hskip 1em plus 0.5em minus
  0.4em\relax Springer, 2014, pp. 536--551.

\bibitem{brachmann2016uncertainty}
E.~Brachmann, F.~Michel, A.~Krull, M.~Ying~Yang, S.~Gumhold, \emph{et~al.},
  ``Uncertainty-driven 6d pose estimation of objects and scenes from a single
  rgb image,'' in \emph{Proceedings of the IEEE Conference on Computer Vision
  and Pattern Recognition}, 2016, pp. 3364--3372.

\bibitem{xiang2017posecnn}
Y.~Xiang, T.~Schmidt, V.~Narayanan, and D.~Fox, ``Posecnn: A convolutional
  neural network for 6d object pose estimation in cluttered scenes,''
  \emph{arXiv preprint arXiv:1711.00199}, 2017.

\bibitem{tekin2018real}
B.~Tekin, S.~N. Sinha, and P.~Fua, ``Real-time seamless single shot 6d object
  pose prediction,'' in \emph{Proceedings of the IEEE Conference on Computer
  Vision and Pattern Recognition}, 2018, pp. 292--301.

\bibitem{peng2019pvnet}
S.~Peng, Y.~Liu, Q.~Huang, X.~Zhou, and H.~Bao, ``Pvnet: Pixel-wise voting
  network for 6dof pose estimation,'' in \emph{Proceedings of the IEEE
  Conference on Computer Vision and Pattern Recognition}, 2019, pp. 4561--4570.

\bibitem{hinterstoisser2012model}
S.~Hinterstoisser, V.~Lepetit, S.~Ilic, S.~Holzer, G.~Bradski, K.~Konolige, and
  N.~Navab, ``Model based training, detection and pose estimation of
  texture-less 3d objects in heavily cluttered scenes,'' in \emph{Asian
  conference on computer vision}.\hskip 1em plus 0.5em minus 0.4em\relax
  Springer, 2012, pp. 548--562.

\bibitem{kehl2017ssd}
W.~Kehl, F.~Manhardt, F.~Tombari, S.~Ilic, and N.~Navab, ``Ssd-6d: Making
  rgb-based 3d detection and 6d pose estimation great again,'' in
  \emph{Proceedings of the IEEE International Conference on Computer Vision},
  2017, pp. 1521--1529.

\bibitem{sundermeyer2018implicit}
M.~Sundermeyer, Z.-C. Marton, M.~Durner, M.~Brucker, and R.~Triebel, ``Implicit
  3d orientation learning for 6d object detection from rgb images,'' in
  \emph{Proceedings of the European Conference on Computer Vision (ECCV)},
  2018, pp. 699--715.

\bibitem{morrison2020GGCNN}
D.~Morrison, P.~Corke, and J.~Leitner, ``Learning robust, real-time, reactive
  robotic grasping,'' \emph{The International Journal of Robotics Research},
  vol.~39, no. 2-3, pp. 183--201, 2020.

\bibitem{zeng2019tossingbot}
A.~Zeng, S.~Song, J.~Lee, A.~Rodriguez, and T.~Funkhouser, ``Tossingbot:
  Learning to throw arbitrary objects with residual physics. arxiv preprint
  arxiv: 190311239,'' 2019.

\bibitem{levine2016end}
S.~Levine, C.~Finn, T.~Darrell, and P.~Abbeel, ``End-to-end training of deep
  visuomotor policies,'' \emph{The Journal of Machine Learning Research},
  vol.~17, no.~1, pp. 1334--1373, 2016.

\bibitem{Levine:2017cu}
S.~Levine, P.~Pastor, A.~Krizhevsky, J.~Ibarz, and D.~Quillen, ``{Learning
  hand-eye coordination for robotic grasping with deep learning and large-scale
  data collection},'' \emph{The International Journal of Robotics Research},
  vol.~37, no. 4-5, pp. 421--436, June 2017.

\bibitem{kalashnikov2018qt}
D.~Kalashnikov, A.~Irpan, P.~Pastor, J.~Ibarz, A.~Herzog, E.~Jang, D.~Quillen,
  E.~Holly, M.~Kalakrishnan, V.~Vanhoucke, \emph{et~al.}, ``Qt-opt: Scalable
  deep reinforcement learning for vision-based robotic manipulation,''
  \emph{arXiv preprint arXiv:1806.10293}, 2018.

\bibitem{Fangyi2018}
\BIBentryALTinterwordspacing
F.~Zhang, J.~Leitner, M.~Milford, and P.~Corke, ``Sim-to-real transfer of
  visuo-motor policies for reaching in clutter: Domain randomization and
  adaptation with modular networks,'' \emph{CoRR}, vol. abs/1709.05746ß, 2017.
  [Online]. Available: \url{http://arxiv.org/abs/1709.05746}
\BIBentrySTDinterwordspacing

\bibitem{redmon2017yolo9000}
J.~Redmon and A.~Farhadi, ``Yolo9000: better, faster, stronger,'' in
  \emph{Proceedings of the IEEE conference on computer vision and pattern
  recognition}, 2017, pp. 7263--7271.

\bibitem{frey2007clustering}
B.~J. Frey and D.~Dueck, ``Clustering by passing messages between data
  points,'' \emph{science}, vol. 315, no. 5814, pp. 972--976, 2007.

\bibitem{redmon2016you}
J.~Redmon, S.~Divvala, R.~Girshick, and A.~Farhadi, ``You only look once:
  Unified, real-time object detection,'' in \emph{Proceedings of the IEEE
  conference on computer vision and pattern recognition}, 2016, pp. 779--788.

\bibitem{He:2016tt}
K.~He, X.~Zhang, S.~Ren, and J.~Sun, ``{Deep Residual Learning for Image
  Recognition},'' pp. 770--778, 2016.

\bibitem{lin2017feature}
T.-Y. Lin, P.~Doll{\'a}r, R.~Girshick, K.~He, B.~Hariharan, and S.~Belongie,
  ``Feature pyramid networks for object detection,'' in \emph{Proceedings of
  the IEEE Conference on Computer Vision and Pattern Recognition}, 2017, pp.
  2117--2125.

\bibitem{Viereck:2017uq}
U.~Viereck, A.~ten Pas, K.~Saenko, and R.~Platt, ``{Learning a visuomotor
  controller for real world robotic grasping using simulated depth images},''
  \emph{arXiv.org}, June 2017.

\bibitem{peng2017sim}
X.~B. Peng, M.~Andrychowicz, W.~Zaremba, and P.~Abbeel, ``Sim-to-real transfer
  of robotic control with dynamics randomization,'' \emph{arXiv preprint
  arXiv:1710.06537}, 2017.

\bibitem{gazebo2004}
N.~P. Koenig and A.~Howard, ``Design and use paradigms for gazebo, an
  open-source multi-robot simulator.'' in \emph{IROS}, vol.~4.\hskip 1em plus
  0.5em minus 0.4em\relax Citeseer, 2004, pp. 2149--2154.

\bibitem{ImageNet}
J.~Deng, W.~Dong, R.~Socher, L.-J. Li, K.~Li, and L.~Fei-Fei, ``{ImageNet: A
  large-scale hierarchical image database},'' in \emph{2009 IEEE Computer
  Society Conference on Computer Vision and Pattern Recognition Workshops (CVPR
  Workshops)}.\hskip 1em plus 0.5em minus 0.4em\relax IEEE, pp. 248--255.

\bibitem{kingma2014adam}
D.~P. Kingma and J.~Ba, ``Adam: A method for stochastic optimization,''
  \emph{arXiv preprint arXiv:1412.6980}, 2014.

\end{thebibliography}

\end{document}